\pdfoutput=1

\documentclass[11pt]{article}

\usepackage[preprint]{acl}

\usepackage{times}
\usepackage{latexsym}

\usepackage[T1]{fontenc}

\usepackage[utf8]{inputenc}

\usepackage{microtype}

\usepackage{inconsolata}

\usepackage{graphicx}
\usepackage{pifont}
\usepackage{amssymb}
\usepackage{booktabs}
\usepackage{tabularx}
\usepackage{multirow}
\usepackage{makecell}
\usepackage{array}

\usepackage{tikz}
\usepackage{xcolor}
\usepackage{listings}
\usepackage[most]{tcolorbox}
\usepackage{cuted}
\usepackage{longtable}
\usepackage{placeins}
\usepackage{float}
\usepackage{stfloats}
\usepackage{xcolor}
\usepackage{pdfpages}

\newcommand{\cmark}{\ding{51}} 
\newcommand{\xmark}{\ding{55}} 
\newcommand{\mmark}{\ensuremath{\sim}} 

\definecolor{gr}{RGB}{00,180,00}
\title{Chart Question Answering from Real-World Analytical Narratives}

\author{
 \textbf{Maeve Hutchinson\textsuperscript{1}},
 \textbf{Radu Jianu\textsuperscript{1}},
 \textbf{Aidan Slingsby\textsuperscript{1}},
 \textbf{Jo Wood\textsuperscript{1}},
 \textbf{Pranava Madhyastha\textsuperscript{1,2}}
\\
\\
 \textsuperscript{1}City St George's, University of London,
 \textsuperscript{2}The Alan Turing Institute
\\
 \small{
   \textbf{Correspondence:}{\{maeve.hutchinson, pranava.madhyastha\}@citystgeorges.ac.uk}
 }
}

\begin{document}
\maketitle
\begin{abstract}
We present a new dataset for chart question answering (CQA) constructed from visualization notebooks. The dataset features real-world, multi-view charts paired with natural language questions grounded in analytical narratives. Unlike prior benchmarks, our data reflects ecologically valid reasoning workflows. Benchmarking state-of-the-art multimodal large language models reveals a significant performance gap, with GPT-4.1 achieving an accuracy of 69.3\%, underscoring the challenges posed by this more authentic CQA setting.

\end{abstract}

\section{Introduction}

Data visualizations are an essential modality for communicating complex information about data. Alongside natural language, they serve as a key medium for communication across domains. As such, the ability to interpret and reason about visualizations is a crucial skill.

As multimodal large language models (MLLMs) evolve beyond simple perception tasks towards becoming visual assistants, there is growing interest in their ability to perform visual reasoning over structured data, including charts and other forms of data visualization. Tasks such as Chart Question Answering (CQA) have emerged for benchmarking a model's visualization reasoning capabilities. 

In this work, we introduce a new dataset for CQA that aims to reflect the complexity of real-world data analysis. \footnote{Dataset available at: \url{https://huggingface.co/datasets/maevehutch/realworld-chartqa}} The dataset is constructed from student authored visualization notebooks, which combine explanatory analytical narrative with custom visualizations. Unlike existing CQA datasets, our dataset is grounded in ecologically valid analytical workflows. To situate this contribution, we first review prior work on visualization literacy and CQA. We then detail our data collection and question generation process, describing the structure and composition of the dataset. Finally, we report some initial benchmarking results using state-of-the-art MLLMs.

\section{Related Work}
\textbf{Visualization Literacy} datasets such as the visualization literacy assessment test (VLAT) \citep{lee_vlat_2017} were initially created to assess human understanding of data visualizations. Recently, they have also been applied to probe the visualization literacy of MLLMs \citep{bendeck_empirical_2024}. These manually curated datasets present small sets of charts paired with multiple-choice questions that probe the ability to perform specific analytic tasks such as retrieving values, identifying trends, or making comparisons. Whilst these tasks seem to mimic real-world analytical workflows \cite{amar_low-level_2005}, the hand-crafted design of these datasets limits their ability to accurately reflect the complexity of real-world visualization reasoning. 

\textbf{Chart Question Answering (CQA)} is the task of answering a natural language question about a visualization image. CQA datasets are designed to benchmark the chart understanding capabilities of models. Early CQA benchmarks such as FigureQA \citep{kahou_figureqa_2018}, DVQA \citep{kafle_dvqa_2018}, and LEAF-QA \citep{chaudhry_leaf-qa_2020} used template-based questions and synthetically generated tasks. Again, these controlled settings are limited.

More recently, CQA datasets have moved toward real-world visualization images. \citet{kim_answering_2020} and ChartQA \citep{masry-etal-2022-chartqa} introduced chart images scraped from real-world reports and online sources. However, these datasets still only have questions that refer to a single chart, and do not include visualizations with multiple views or interactive elements. These datasets begin to reflect more realistic evaluation settings, but still do not completely capture visualization as done in-practice, where users often engage with visualizations that have multiple views, such as dashboards or linked visualizations.

Some newer datasets begin to address this. CharXiv \citep{wang_charxiv_2024} includes charts composed of multiple sub views, although its questions still focus on one image. MultiChartQA \citep{zhu-etal-2025-multichartqa} allows questions to target multiple related visualizations, moving closer to the kinds of cross-chart reasoning analysts perform in practice. However, these datasets are still composed solely of static visualizations.

Another important distinction lies in how questions are generated. Some datasets, such as VLAT \citep{lee_vlat_2017} and MultiChartQA \citep{zhu-etal-2025-multichartqa}, rely exclusively on human-authored questions. While this approach ensures high-quality queries aligned with human reasoning, the scalability of dataset construction is limited. Conversely, other datasets like ChartQA \citep{masry-etal-2022-chartqa} and CharXiv \citep{wang_charxiv_2024} adopt semi-automatic approaches, using models to produce questions alongside human validation, enabling larger datasets across more images.

Notably, previous datasets, whether template, human or machine-authored, are generated from the visualization image, caption, or from post hoc chart summaries. This often as a result of data collection processes that extract chart images in isolation, often scraped from online sources, removed from the surrounding analytical narrative. Due to the nature of source materials, this analytical context often does not exist at all and is left entirely implicit, available only from the visual context. The nature of these online sources may also raise copyright concerns due to the use of third-party images without explicit permission.

\section{Methods}
\subsection{Data Collection}
Our dataset is derived from literate visualization (litvis) notebooks, structured markdown documents that combine narrative analysis, code, embedded datasets, and inline visualizations \cite{wood_design_2019}. The notebooks were authored by undergraduate and postgraduate students as part of their final coursework for a 10-week data visualization module. These notebooks offer an ecologically valid window into real-world analytical practice: students independently selected datasets to analyze, posed research questions, and designed custom visualizations to explore those questions. These notebooks surface articulations of analytical reasoning that are typically left implicit in other sources of visualizations, providing a rich basis for question generation. See appendix~\ref{sec:litvis} for an example notebook.

We applied several filtering steps to ensure data quality. Submissions were excluded if they lacked visualizations, included personally identifiable information, lacked sufficient narrative, or otherwise failed to meet basic quality thresholds. After filtering, we retained 22 notebooks for further processing.

From each retained notebook, we extracted two primary sources of data: the analytical narrative written by the student, and the corresponding visualizations.  Visualizations were captured by rendering each notebook in HTML and using a headless browser to take screenshots of the embedded figures. Interactive visualizations were present in many of the notebooks, a feature missing from many sources of visualizations in CQA. To partially capture these interactive dynamics, we developed a method for capturing some interactive views statically. For visualizations with discrete interactive controls, such as radio buttons or drop-down menus, we systematically enumerated all categorical options and recorded screenshots of each interactive view. This allowed us to collect multiple views of the same visualization, reflecting user-driven analytical actions that are absent in existing datasets. To prepare the narrative for question generation, we segmented the extracted content into chunks of at most 200 words.

\begin{table*}[ht!]
\centering
\resizebox{0.9\textwidth}{!}{
\begin{tabular}{@{}r|ccc|cc@{}}
\toprule
\textbf{Dataset} & 
\multicolumn{3}{c|}{\textbf{Visualizations}} & 
\multicolumn{2}{c}{\textbf{Questions}} \\
\cmidrule(lr){2-4} \cmidrule(lr){5-6}
& Real-World & \# Chart Types & Multi/Interactive & Unanswerable & Narrative Context \\
\midrule
LeafQA \citeyearpar{chaudhry_leaf-qa_2020} & \xmark & 6           & \xmark/\xmark & \xmark & \xmark \\
\citet{kim_answering_2020}  & \mmark & 2           & \xmark/\xmark & \xmark & \xmark \\
ChartQA \citeyearpar{masry-etal-2022-chartqa}      & \cmark & 3           & \xmark/\xmark & \xmark & \xmark \\
CharXiv \citeyearpar{wang_charxiv_2024}    & \cmark & \textit{unbounded} & \cmark/\xmark & \cmark & \xmark \\
MultiChartQA \citeyearpar{zhu-etal-2025-multichartqa} & \cmark & \textit{unbounded} & \cmark/\xmark & \cmark & \xmark \\
\textbf{Ours} & \cmark & \textit{unbounded} & \cmark/\cmark & \cmark & \cmark \\
\bottomrule
\end{tabular}
}
\caption{Comparison between our dataset and existing chart question-answering datasets, grouped by visualization and question characteristics.}
\label{tab:comparison}
\end{table*}

\subsection{Question Generation}
We structured our dataset according to established analytical task taxonomies from visualization research to ensure that the questions in our dataset reflect realistic analytical goals. Specifically, we adopt the eight task categories defined in the VLAT \cite{lee_vlat_2017}, which were curated from prior task taxonomies by \citet{amar_low-level_2005} and \citet{chen_toward_2009}. These tasks are: Retrieve Value, Find Extremum, Find Correlations, Make Comparisons, Characterize Distribution, Determine Range, Find Anomalies, and Find Clusters.

Our question generation pipeline centers on the analytical narrative authored by students. This approach is inspired by \citeposs{changpinyo-etal-2022-may} work in visual question answering (VQA), who demonstrate the viability of generating high-quality question-answer pairs from language context rather than visual context. This approach allows us to generate meaningful, grounded questions using an LLM without parsing the chart images.

For each segment, we prompted an LLM to generate a question-answer pair grounded in the context. The prompt provided a short description of each task category with representative examples. The model was asked to extract a relevant quote from the narrative, use it to generate a question-answer pair, and classify the pair according to the task taxonomy. The quote extraction allows us to verify the fidelity of the pair later in our validation process. 

We then prompted the LLM to generate multiple choice distractors. The model received the narrative context, question-answer pair, and task classification, and was instructed to generate three plausible but incorrect alternative answers. The distractors were designed to match the structure and domain of the correct answer. Additionally, we appended a fifth answer option: \textit{"Cannot be determined from the visualization(s)"}. This serves both as a realistic distractor and also as a correct answer choice for some questions, which will be determined during the validation process. Full prompt templates are provided in appendix ~\ref{sec:prompts}.

This pipeline yielded an initial set of 429 multiple-choice QA pairs, each grounded in the analytical context and aligned to an analytical task. These pairs then underwent a rigorous manual validation process.

\subsection{Human Validation}

All 429 LLM-generated QA pairs underwent stringent human validation by a data visualization expert to ensure the quality and reliability of the dataset. Each pair was reviewed against a set of rejection criteria, targeting two primary sources of invalid questions: (1) misalignment with the available visualizations, and (2) quality issues arising from the narrative context or generation process.

The first criterion focused on visualization alignment. Some visualizations were unable to render due to the unavailability of the underlying datasets, and because our QA generation process operated on the narrative context alone, some generated pairs referred to visualizations that could not be recovered during our data collection pipeline. Any QA pair that could not be reliably related to at least one available visualization was excluded.

The second rejection criterion addressed the scope of the narrative context and generation quality. Some students describe aspects unrelated to analytical insights, such as dataset collection challenges, findings they found surprising, or general reflections. While these are interesting and valuable parts of the students' process, they are out of scope for this dataset and so QA pairs generated from this context were excluded.

During validation, we also explicitly associated each accepted QA pair with the specific views it referenced, as each notebook often included multiple charts. In some cases, questions required information that was only visible interactive views not captured, often tooltip values. When a question did relate to an available chart but remained unanswerable due to missing context, we retained it and assigned it "cannot be determined".


\begin{table*}[ht!]
\centering
\resizebox{0.8\textwidth}{!}{
\begin{tabular}{@{}r c|ccc@{}}
\toprule
\textbf{Task} & \textbf{Count} & \textbf{GPT-4.1} & \textbf{Qwen2.5-VL-32B} & \textbf{Qwen2.5-VL-7B} \\
\midrule
All                     & 205 & 69.27\% & 56.59\% & 31.71\% \\
\midrule
Retrieve Value            & 68 & 76.47\% & 55.88\% & 25.00\% \\
Find Extremum             & 55 & 69.09\% & 60.00\% & 36.36\% \\
Find Correlations         & 22 & 72.73\% & 54.55\% & 27.27\% \\
Make Comparisons          & 22 & 50.00\% & 59.09\% & 50.00\% \\
Characterize Distribution & 15 & 66.67\% & 46.67\% & 20.00\% \\
Determine Range           & 12 & 75.00\% & 58.33\% & 41.67\% \\
Find Anomalies            & 9  & 44.44\% & 55.56\% & 33.33\% \\
Find Clusters             & 2  & 100.00\% & 50.00\% & 0.00\% \\
\bottomrule
\end{tabular}
}
\caption{Accuracy by task type for GPT-4.1 and Qwen2.5-VL models. The top row reports overall accuracy across all tasks, followed a task breakdown, ordered by task frequency.}
\label{tab:model-accuracy}
\end{table*}

\section{Dataset Analysis}

Following validation, we retained 205 high-quality QA pairs, corresponding to 103 visualization images. 75 questions, 36.6\%, have multiple visualization images or multiple views. 33 questions, 16.1\% of questions are unanswerable. Table~\ref{tab:comparison} provides a comparison of our dataset to previous work across key visualization and question characteristics.


Table~\ref{tab:model-accuracy} provides a breakdown of question types in the dataset by visualization task. The observed imbalance reflects the natural distribution of analytical strategies employed by students in their projects. Tasks such as Retrieve Value and Find Extremum are most common, suggesting a strong emphasis on identifying specific data points or extreme values. Conversely, higher-order tasks like Find Clusters or Find Anomalies are relatively rare.

\section{Model Evaluation}


We evaluated the performance of two state-of-the-art vision-language models on our dataset: OpenAI’s proprietary GPT-4.1 \citep{openai_gpt-41_2025} and Alibaba’s open-weight Qwen2.5-VL models at two parameter scales (7B and 32B) \citep{bai_qwen25-vl_2025}. Each model was presented with the question and corresponding visualization(s) and tasked with selecting the correct answer from the five multiple-choice options.


As shown in Table~\ref{tab:model-accuracy}, GPT-4.1 achieved the highest accuracy at 69.27\%, outperforming both versions of Qwen2.5-VL. The 32B variant of Qwen2.5-VL attained a moderate accuracy of 56.59\%, while the 7B variant lagged significantly at 31.71\%. This performance disparity underscores the impact of model scale on complex visual question answering tasks. Appendix~\ref{sec:examples} provides some examples from our dataset alongside GPT4.1's responses.


Table~\ref{tab:model-accuracy} presents model accuracy broken down by question type. GPT-4.1 demonstrates consistently strong performance across most tasks, exceeding 66\% accuracy in five of the eight categories. It performs particularly well on Retrieve Value and Determine Range, tasks that rely on precise visual extraction, suggesting strong literal comprehension of chart elements. However, its performance drops on more interpretive tasks such as Make Comparisons (50.00\%), perhaps indicating challenges with contextual or higher-order reasoning. Interestingly, Qwen2.5-VL-32B outperforms GPT-4.1 on these two tasks, despite trailing on most others, suggesting possible strengths in certain visual discrimination tasks. The 7B variant of Qwen2.5-VL performs substantially worse across nearly all categories, aside from Make Comparisons, where it matches GPT-4.1’s performance.

Caution is however warranted when interpreting results for less frequent task types such as Find Anomalies and Find Clusters, which contain relatively few questions. Despite this, the overall trends suggest that performance differences across task types are meaningful, and that structured taxonomies offer useful insight into the capabilities and limitations of current MLLMs in chart understanding.

\section{Conclusion}
Our dataset introduces a more realistic and ecologically grounded benchmark for chart question answering, reflecting how visualizations are created and interpreted in practice. By capturing analytical narratives, multiple and interactive views, it challenges current models in ways prior datasets do not. Initial evaluations highlight substantial performance gaps, pointing to the need for models with deeper reasoning and contextual understanding of visual data. We observe significant variance in model performance across task types, suggesting that certain forms of visual reasoning remain especially challenging. We hope this dataset fosters future research toward more capable and context-aware multimodal systems.

\newpage
\section*{Ethics Statement}
This study and its data collection procedures were formally approved by our university’s Research Ethics Committee. Upon receiving approval, we contacted graduates of the program to inform them about the study’s aims and potential contributions. We obtained explicit informed consent from those who agreed to participate, specifically for the use of their coursework in our research. The dataset exclusively comprises submissions from students who voluntarily provided permission for their materials to be processed and released as part of this research.

\section*{Limitations}
While our dataset offers a more ecologically grounded benchmark for CQA, it has several limitations. Firstly, the task distribution is imbalanced, with lower-level tasks like Retrieve Value more common and higher-order tasks like Find Clusters underrepresented. Future work could curate a more balanced set to cover a wider range of reasoning types. Secondly, the dataset includes only 205 validated question–answer pairs. This limited size reflects our emphasis on rigorous human validation to ensure alignment between questions, narratives, and visualizations. Our methodology could be extended to larger corpora of visualization notebooks to create a more expansive dataset. Finally, all questions are in English. While the tasks are conceptually broad, some formulations may not generalize well across languages. Future efforts could explore multilingual extensions by incorporating narratives from other languages.

\bibliography{acl_latex}

\clearpage
\appendix

\section{Task Information}
\label{sec:task}
\begin{center}
\centering
\small
\begin{tabular}{@{}p{4cm}p{4.5cm}p{6.5cm}@{}}
\toprule
\textbf{Task Name \& Description} & \textbf{Pro Forma Abstract} & \textbf{Examples (Q → A)} \\
\midrule
\textbf{Retrieve Value} \newline Given a set of specific cases, find attributes of those cases. & What are the values of \newline attributes \{X, Y, Z, ...\} in the data cases \{A, B, C, ...\}? & 
What was the price of a barrel of oil in February 2015? → \$50 \\
& & What is the average internet speed in Japan? → 15.3 Mbps \\\\
& & What is the weight of the person who is 165.1 cm tall? → 60 kg \\
\midrule
\textbf{Find Extremum} \newline Find data cases possessing an extreme value of an attribute. & What are the top/bottom N data cases with respect to \newline attribute A? & 
In which month was the price of a barrel of oil the lowest in 2015? → August \\
& & Which country has the fastest average internet speed in Asia? → South Korea \\\\
& & What is the height of the tallest person among the 85 males? → 198 cm \\
\midrule
\textbf{Determine Range} \newline Find the span of values of an attribute within a set. & What is the range of values of attribute A in a set S of data cases? & 
What was the price range of a barrel of oil in 2015? → \$38 to \$60 \\
& & What is the range of average internet speeds in Asia? → 4.3 Mbps to 15.3 Mbps \\\\
& & What is the weight range among the 85 males? → 52 kg to 90 kg \\
\midrule
\textbf{Characterize Distribution} \newline Characterize the distribution of a quantitative attribute. & What is the distribution of values of attribute A in a set S of data cases? & 
How is the distribution of taxi passenger ratings characterized? → Skewed to the left \\
& & What is the distribution pattern of student grades in the dataset? → Approximately normal distribution centered around 75\% \\
\midrule
\textbf{Find Anomalies} \newline Identify anomalies within a set of data cases. & Which data cases in a set S of data cases have unexpected/exceptional values? & 
Which individual's height deviates most from the others? → 210 cm \\\\
& & Which city's metro system deviates most from the trend? → Beijing \\
\midrule
\textbf{Find Clusters} \newline Find clusters of similar attribute values. & Which data cases are similar in value for attributes \{X, Y, Z, …\}? & 
Describe any groups of individuals who share similar height and weight characteristics. → A group is clustered around 176 cm in height and 70 kg in weight. \\\\
& & What patterns of similarity can you find among metro systems based on number of stations and system length? → Several metro systems are clustered around 300 stations and 200 km length. \\
\midrule
\textbf{Find Correlations} \newline Determine relationships between two attributes. & What is the correlation between attributes X and Y in a set S? & 
What is the relationship between height and weight? → Negative linear \\
& & How does ridership relate to stations? → Positive correlation \\\\
& & Trend in coffee prices over 2013? → Increasing \\
\midrule
\textbf{Make Comparisons} \newline Compare sets of cases with respect to an attribute. & How do data cases compare with respect to attribute A? & 
Apple vs Huawei market share? → Apple's is larger \\
& & Ratings between 4.6–4.8 and 4.2–4.4? → 4.6–4.8 has more \\\\
& & Shanghai vs Beijing ridership? → Shanghai's is higher \\
\bottomrule
\end{tabular}
\label{tab:task-taxonomy}
\end{center}

\clearpage
\section{Prompts}
\label{sec:prompts}
\begin{center}
\centering
\begin{tcolorbox}[
    title={Prompt: QA Generation},
    fonttitle=\bfseries,
    colback=white,
    colframe=black,
    boxrule=0.5pt,
    sharp corners,
    width=\textwidth, 
    arc=0mm,
    top=1mm,
    bottom=1mm,
    left=2mm,
    right=2mm,
    enhanced,
    breakable
]
\footnotesize
\begin{verbatim}
You are a data visualization expert and question-generation assistant.

Given the following TEXT:

{ANALYTICAL CONTEXT}

Your task is to generate between 3 and 10 QUESTION-ANSWER pairs based on the TEXT, 
and assign each one to the most appropriate TASK listed below.

Only generate questions if the information in the TEXT is clearly related to a task.

{TASK INFORMATION}

### Output Instructions:
- For each QA pair, include:
    - The direct **quote** from the TEXT
    - The **question**
    - The **answer**, which should be concise and suitable for a multiple choice test
    - The **most appropriate TASK** name from the list
- Only generate a question if it fits into one of the tasks.
- Do not repeat questions
- Prefer fewer, high-quality questions
- Avoid yes/no or true/false answers.
- Output must be a JSON list of dictionaries, like this:

```json
[
  {"quote": "Example quote", "q": "Example question?", "a": "Answer.", "task": "Retrieve Value"},
  ...
]
```
\end{verbatim}
\end{tcolorbox}
\end{center}

\begin{center}
\centering
\begin{tcolorbox}[
    title={Prompt: Answer Choices Generation},
    fonttitle=\bfseries,
    colback=white,
    colframe=black,
    boxrule=0.5pt,
    sharp corners,
    width=\textwidth, 
    arc=0mm,
    top=1mm,
    bottom=1mm,
    left=2mm,
    right=2mm,
    enhanced,
    breakable
]
\footnotesize
\begin{verbatim}
You are creating a multiple choice question about data visualization.

Given the following context:
Context: {ANALYTICAl CONTEXT}

We have a question and answer pair:
Question: {QUESTION}
Correct Answer: {ANSWER}

Generate 3 **plausible but incorrect** answer choices. These should:
- Be related to the same context
- Be in the same format as the correct answer 
(e.g. numerical with the same units, textual with similar length)
- Be different from the correct answer
- Be wrong
- DO NOT make answers that are along the lines of cannot be determined/don't know/can't tell

Output as only a Python list: ["a1", "a2", a3"]
\end{verbatim}
\end{tcolorbox}
\end{center}

\begin{center}
\centering
\begin{tcolorbox}[
    title={Prompt: Model Evaluation},
    fonttitle=\bfseries,
    colback=white,
    colframe=black,
    boxrule=0.5pt,
    sharp corners,
    width=\textwidth, 
    arc=0mm,
    top=1mm,
    bottom=1mm,
    left=2mm,
    right=2mm,
    enhanced,
    breakable
]
\footnotesize
\begin{verbatim}
Question: {QUESTION}

Answer choices: {ANSWER CHOICES}

Please respond with ONLY the letter (A, B, C, D or E) corresponding to your answer.
\end{verbatim}
\end{tcolorbox}
\end{center}

\clearpage
\section{Examples from the Dataset}
\label{sec:examples}
\begin{center}
\begin{tcolorbox}[
    title={Faceted Views},
    fonttitle=\bfseries,
    colback=white,
    colframe=black,
    boxrule=0.5pt,
    sharp corners,
    width=\textwidth, 
    arc=0mm,
    top=1mm,
    bottom=1mm,
    left=2mm,
    right=2mm,
    enhanced,
    breakable
]
    \begin{center}
    \includegraphics[width=0.5\textwidth]{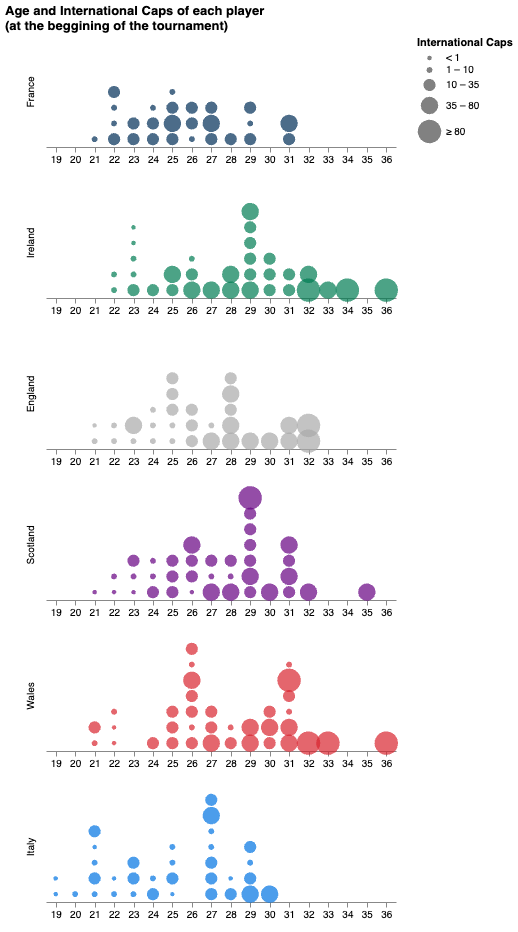}
    \end{center}
    
    \vspace{0.5cm}
    \textbf{Retrieve Value:} What is the range of ages in the France rugby team?\\
    \textbf{Answers:} 14 years,  \textcolor{gr}{10 years}, 8 years, 15 years, Cannot be determined from the visualization(s)\\
    \textbf{GPT 4.1:} \textcolor{red}{8 years}
    
    \vspace{0.5cm}
    \textbf{Find Extremum:} Which team has the narrowest age range?\\
    \textbf{Answers:} \textcolor{gr}{France}, Ireland, Scotland, Wales, Cannot be determined from the visualization(s)]\\
    \textbf{GPT 4.1:} \textcolor{gr}{France}

\vspace{0.5cm}
\textbf{Make Comparisons:} How does the age range of the France rugby team compare to that of Wales?\\
\textbf{Answers:} France's range is wider than Wales', France's range is the same as Wales', \textcolor{gr}{France's range is narrower than Wales'}, France's range is 7 years less than Wales', Cannot be determined from the visualization(s)\\
\textbf{GPT 4.1:} \textcolor{gr}{France's range is narrower than Wales'}
\end{tcolorbox}
\end{center}

\clearpage
\begin{center}
\begin{tcolorbox}[
    title={Multiple Images},
    fonttitle=\bfseries,
    colback=white,
    colframe=black,
    boxrule=0.5pt,
    sharp corners,
    width=\textwidth, 
    arc=0mm,
    top=1mm,
    bottom=1mm,
    left=2mm,
    right=2mm,
    enhanced,
    breakable
]
    \begin{center}
    \includegraphics[width=0.3\textwidth]{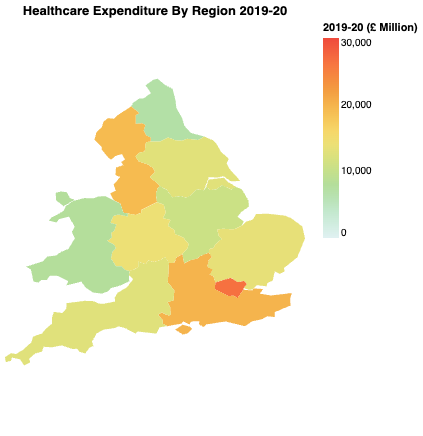}
    \includegraphics[width=0.4\textwidth]{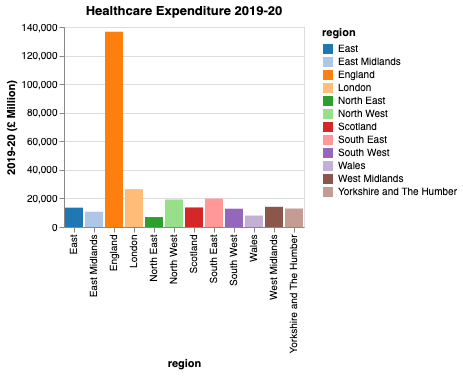}
    \includegraphics[width=0.15\textwidth]{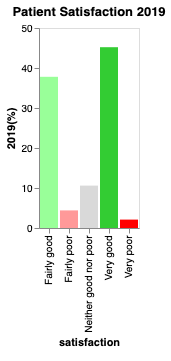}
    
    \vspace{0.2cm}

    \includegraphics[width=0.3\textwidth]{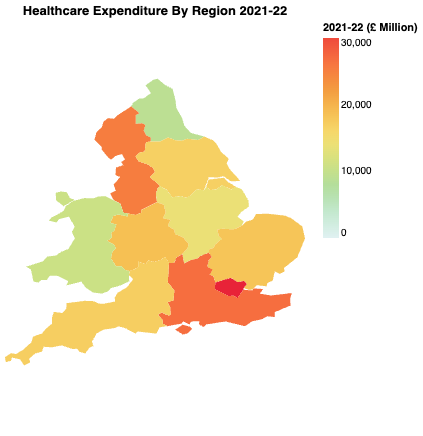}
    \includegraphics[width=0.4\textwidth]{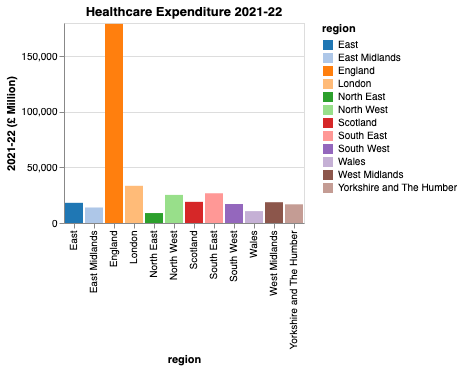}
    \includegraphics[width=0.15\textwidth]{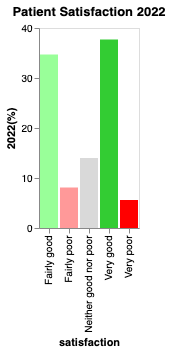}
    \hspace{0.2cm}

    \end{center}
    
\vspace{0.5cm}
\textbf{Find Correlations:} What is the relationship between healthcare expenditure and patient satisfaction between 2019 and 2022?\\
\textbf{Answers:} Patient satisfaction remained relatively stable despite increased expenditure., Healthcare expenditure declined, leading to decreased patient satisfaction., Patient satisfaction increased with increased expenditure., \textcolor{gr}{Despite increased expenditure, patient satisfaction declined.}, Cannot be determined from the visualization(s)\\
\textbf{GPT 4.1:} \textcolor{red}{Patient satisfaction remained relatively stable despite increased expenditure.}

\end{tcolorbox}
\end{center}

\begin{center}
\begin{tcolorbox}[
    title={Multiple Images},
    fonttitle=\bfseries,
    colback=white,
    colframe=black,
    boxrule=0.5pt,
    sharp corners,
    width=\textwidth, 
    arc=0mm,
    top=1mm,
    bottom=1mm,
    left=2mm,
    right=2mm,
    enhanced,
    breakable
]
    \begin{center}
    \includegraphics[width=0.3\textwidth]{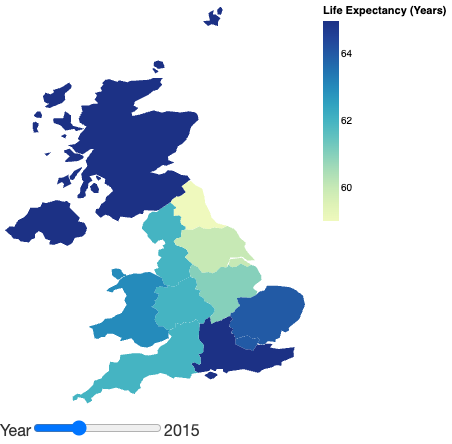}
    \includegraphics[width=0.3\textwidth]{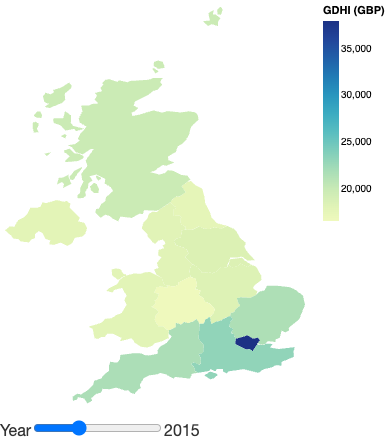}
    \end{center}
    \vspace{0.5cm}
\textbf{Retrieve Value:} What is the life expectancy and GDHI of Northern Ireland?\\
\textbf{Answers:} 65 years and £20,916, \textcolor{gr}{65 years and £17,916}, 60 years and £27,916, 75 years and £15,916, Cannot be determined from the visualization(s)\\
\textbf{GPT 4.1:} \textcolor{red}{65 years and £20,916}

\end{tcolorbox}
\end{center}

\clearpage
\begin{center}
\begin{tcolorbox}[
    title={Interactive View, Cannot be determined},
    fonttitle=\bfseries,
    colback=white,
    colframe=black,
    boxrule=0.5pt,
    sharp corners,
    width=\textwidth, 
    arc=0mm,
    top=1mm,
    bottom=1mm,
    left=2mm,
    right=2mm,
    enhanced,
    breakable
]
    \begin{center}
    \includegraphics[width=0.7\textwidth]{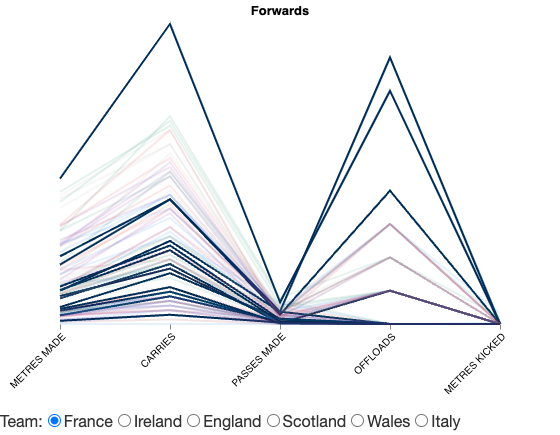}
    \end{center}
    
    \vspace{0.5cm}
\textbf{Find Anomalies:} Which French forwards have unusually high offload numbers compared to other forwards?\\
\textbf{Answers:} Gael Fickou and Damian Penaud, Gregory Alldritt and Antoine Dupont, Cyril Baille and Francois Cros, Cyril Baille and Gregory Alldritt, \textcolor{gr}{Cannot be determined from the visualization(s)}\\
\textbf{GPT 4.1:} \textcolor{red}{Cyril Baille and Gregory Alldritt}
\end{tcolorbox}
\end{center}

\section{Example Literate Visualization Notebook}
\label{sec:litvis}
\includegraphics[width=0.95\textwidth]{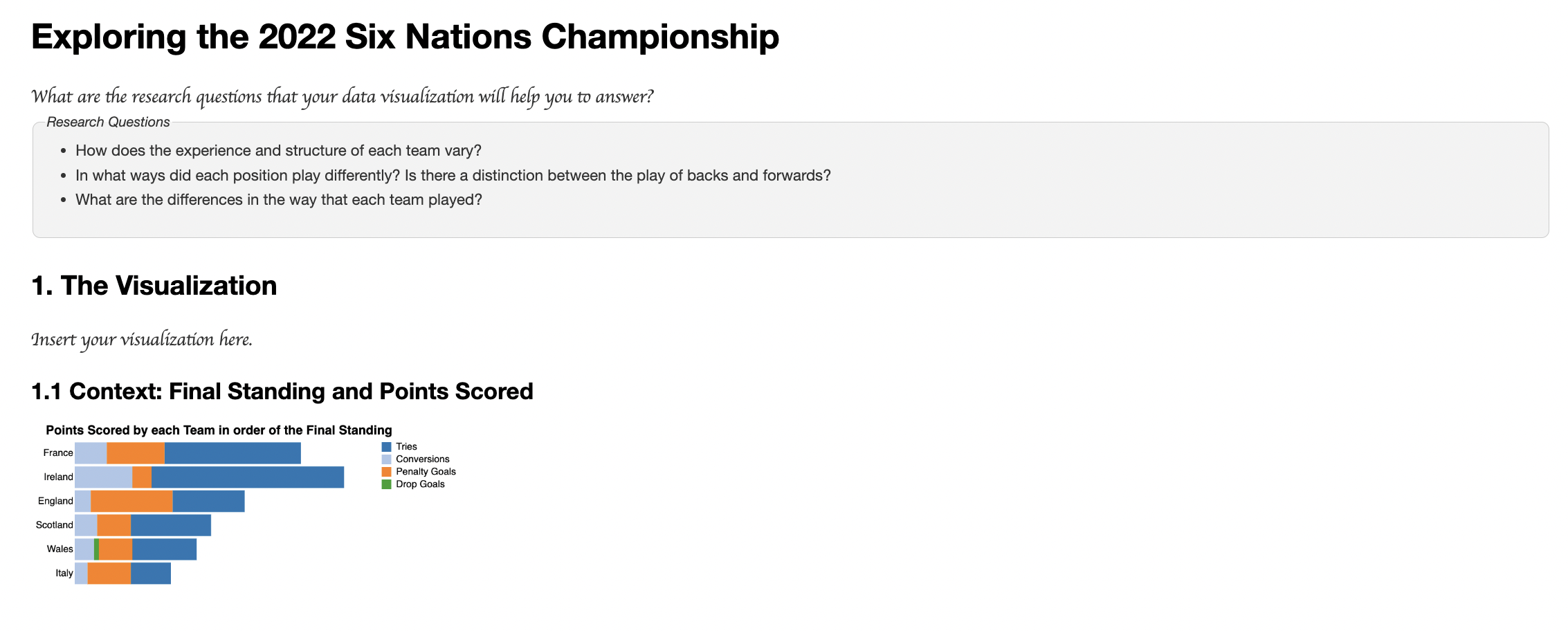}\par\vspace{1em}

\clearpage
\includegraphics[width=1\textwidth]{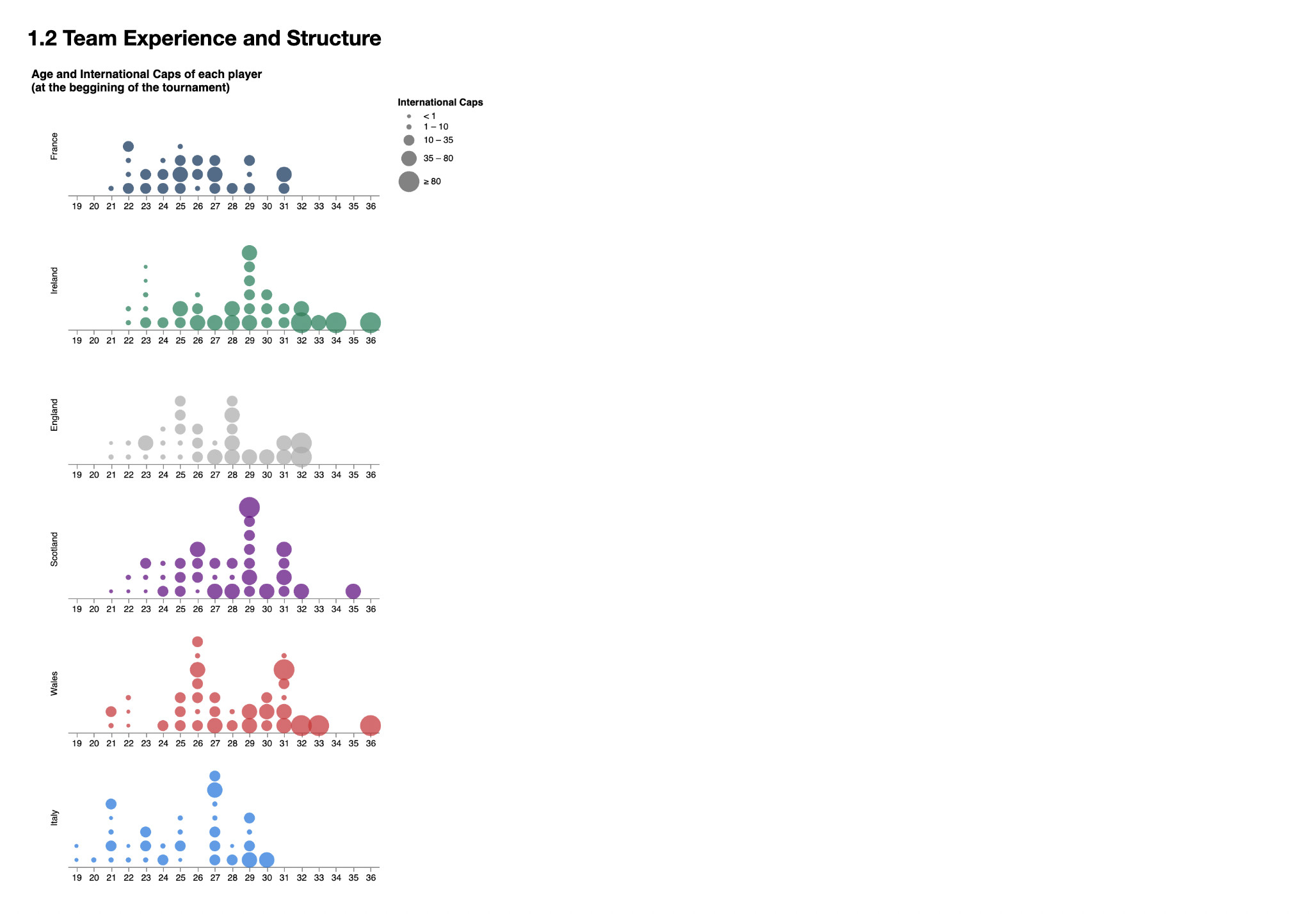}\vspace{0.5cm}

\includegraphics[width=1\textwidth]{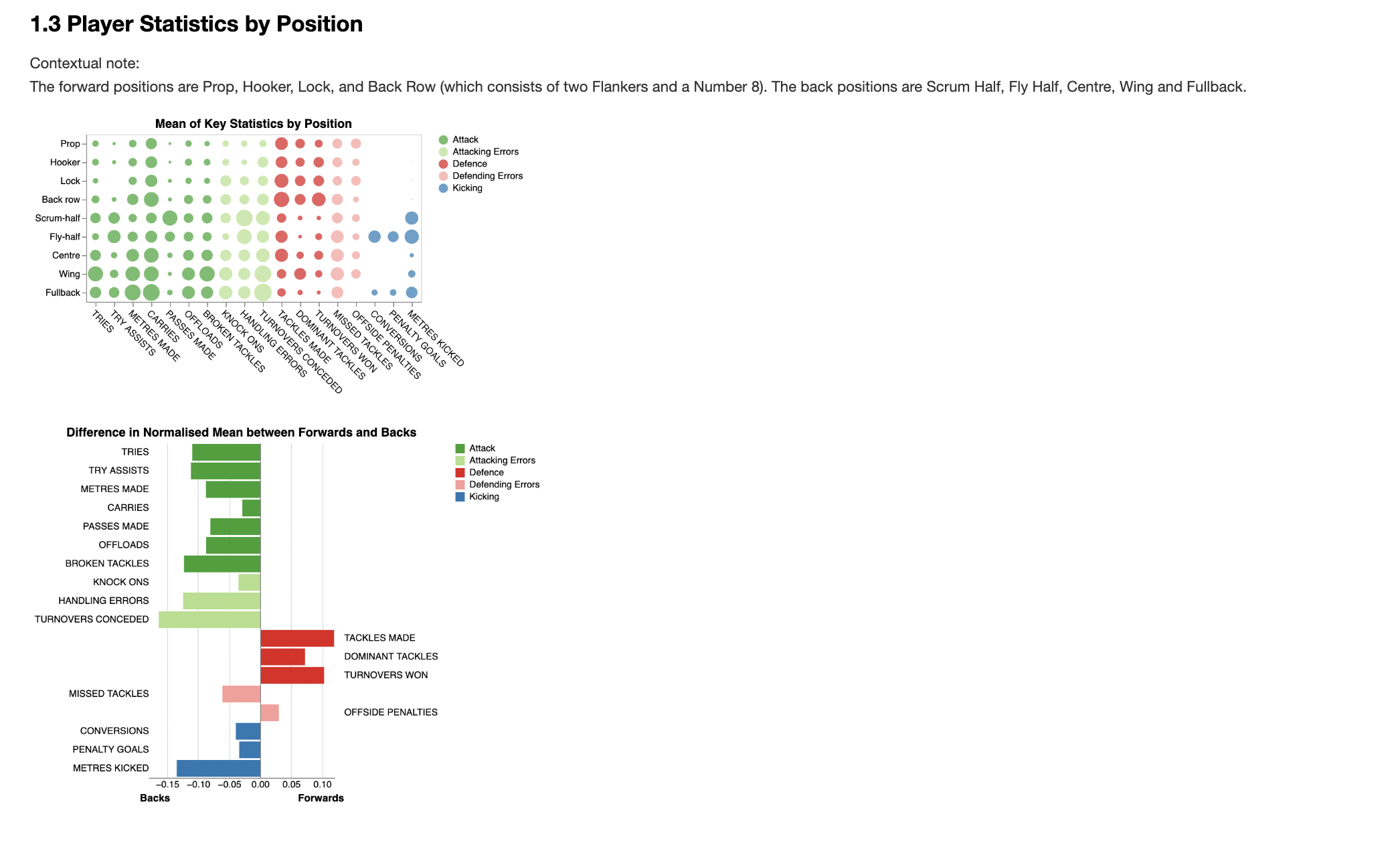}
\clearpage
\includegraphics[width=1\textwidth]{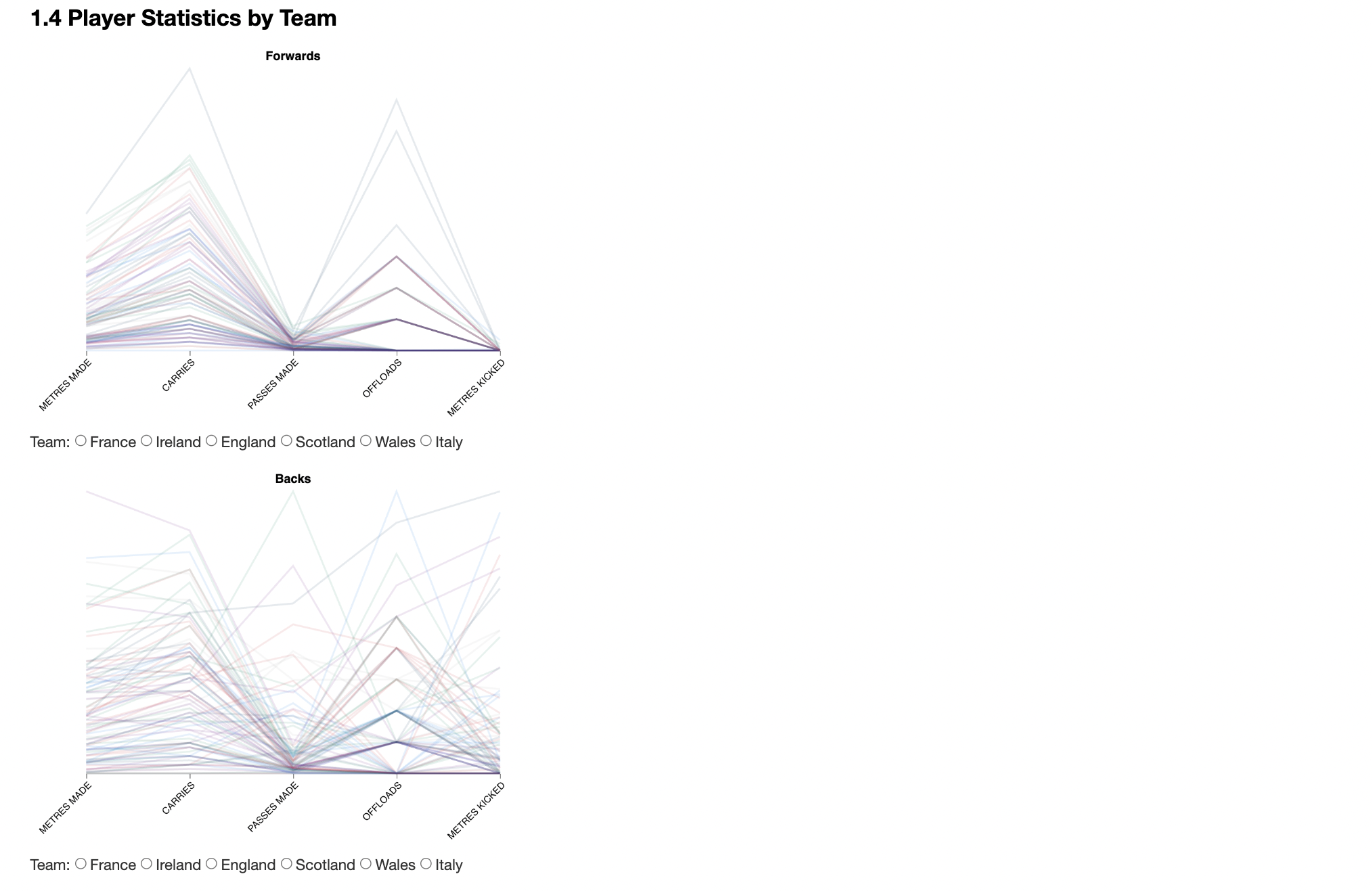}\par\vspace{0.5cm}

\includegraphics[width=1\textwidth]{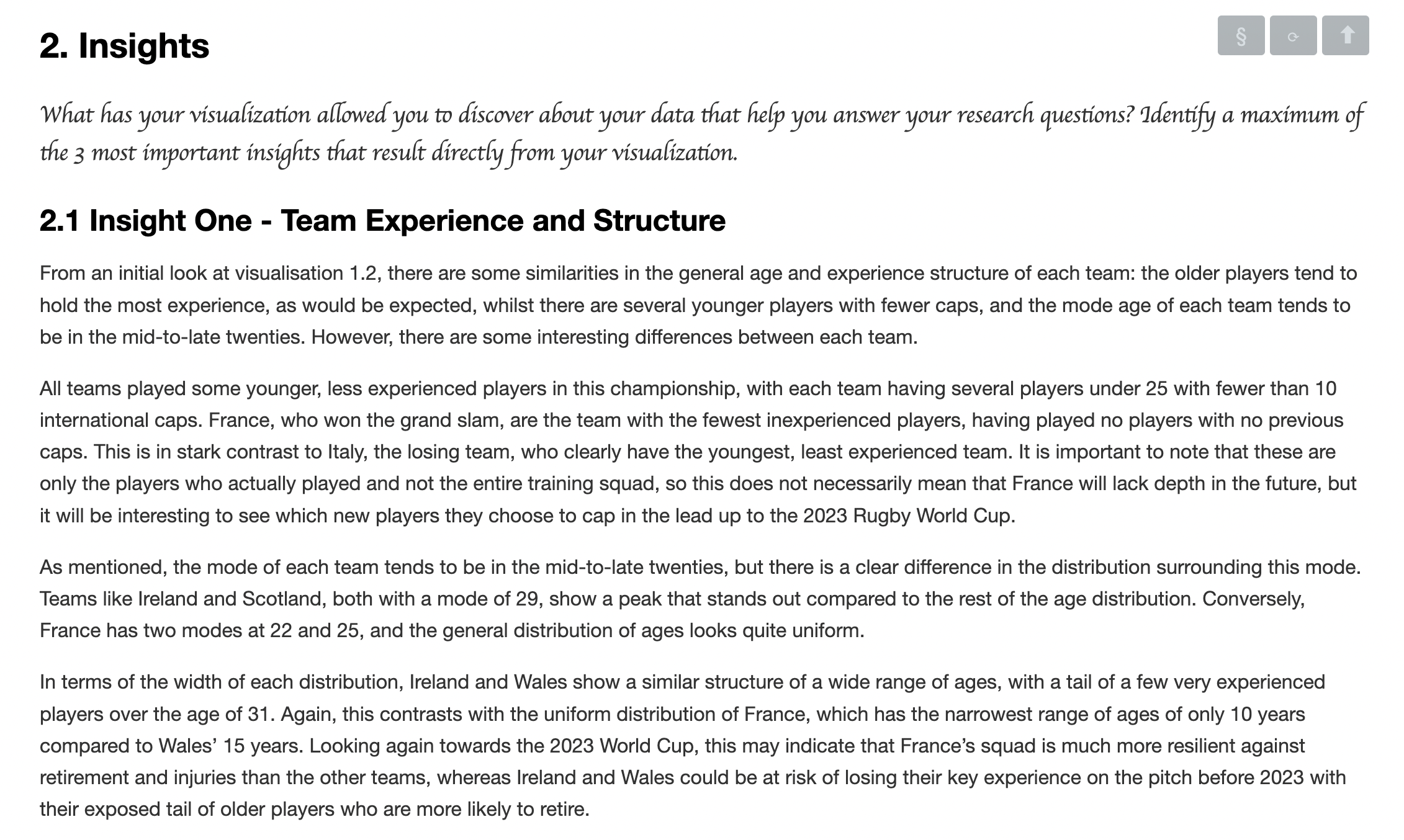}\par

\clearpage
\includegraphics[width=1\textwidth]{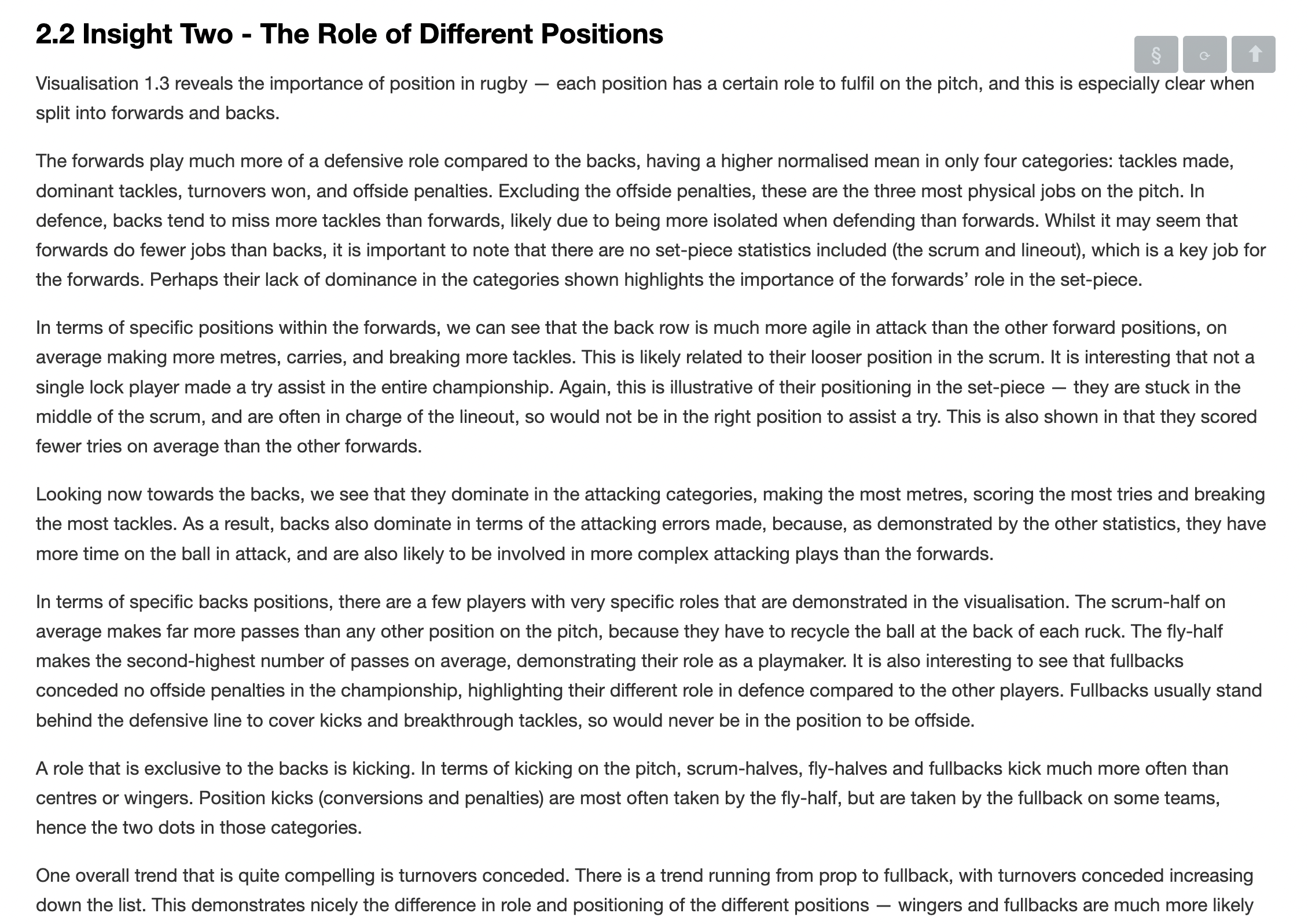}\par
\includegraphics[width=1\textwidth]{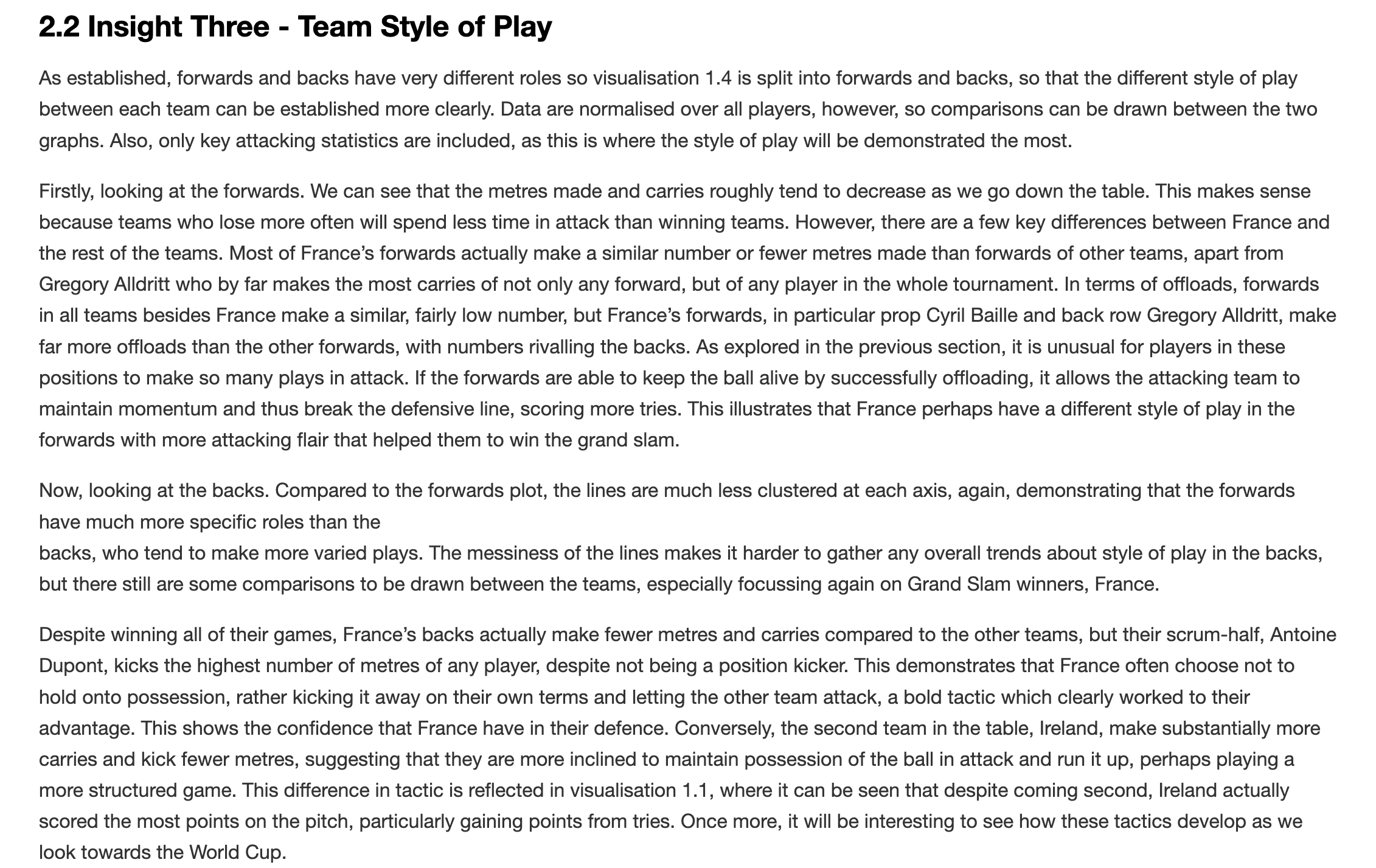}

\end{document}